\documentclass[11pt,english]{article}
\usepackage[latin9]{inputenc}
\usepackage[a4paper]{geometry}
\usepackage{verbatim}
\usepackage{amsmath}
\usepackage{amssymb}
\usepackage{graphicx}
\usepackage{esint}

\makeatletter
\usepackage{nips14submit_e, times}
\usepackage{amsmath}
\usepackage{amsfonts}
\usepackage{amsthm}
\usepackage{enumerate}
    \usepackage[ plainpages = true, pdfpagelabels, 
                 pdfpagelayout = useoutlines,
                 bookmarks,
                 bookmarksopen = true,
                 bookmarksnumbered = true,
                 breaklinks = true,
                 linktocpage,
                 pagebackref,
                 colorlinks = true,
                 linkcolor = blue,
                 urlcolor  = blue,
                 citecolor = red,
                 anchorcolor = green,
                 hyperindex = true,      
                 ]{hyperref} 
\usepackage[round]{natbib}
\usepackage{soul}

\usepackage{url}

\newcommand{\trace}{\mathop{\mathrm{tr}}}

\author{ 
Wittawat Jitkrittum \hspace{6mm} Arthur Gretton \hspace{6mm} Nicolas Heess \thanks{Nicolas Heess is now affiliated with Google Deepmind.} \\
\vspace{2mm}
Gatsby Unit, University College London \\ 
\texttt{ wittawat@gatsby.ucl.ac.uk} \\ 
\texttt{ arthur.gretton@gmail.com } \\ 
\texttt{nheess@gmail.com} 
}

\nipsfinalcopy

\makeatother

\usepackage{babel}
\begin{document}

\title{Passing Expectation Propagation Messages with Kernel Methods}
\maketitle
\begin{abstract}
We propose to learn a kernel-based message operator which takes as
input all expectation propagation (EP) incoming messages to a factor
node and produces an outgoing message. In ordinary EP, computing an
outgoing message involves estimating a multivariate integral which
may not have an analytic expression. Learning such an operator allows
one to bypass the expensive computation of the integral during inference
by directly mapping all incoming messages into an outgoing message.
The operator can be learned from training data (examples of input
and output messages) which allows automated inference to be made on
any kind of factor that can be sampled. 

\begin{comment}
Learning of such operator is done offline with the aid of importance
sampling for computing ground truth projected output messages. 
\end{comment}

\end{abstract}

\section{Background }

Existing approaches to automated probabilistic inference can be broadly
divided into two categories \citep{Heess2013}: uninformed and informed
cases. In the uninformed case, the modeler has full freedom in expressing
a probabilistic model without any constraint on the set of usable
factors. This high flexibility comes at a price during inference as
less factor-specific information is available to the inference engine.
Often MCMC-based sampling techniques are employed by treating the
factors as black boxes. In the informed case \citep{SDT2014,Minka2012},
the modeler is required to build a model from constructs whose necessary
computations are known to the inference engine. Although efficient
during the inference, using an unsupported construct would require
manual derivation and implementation of the relevant computations
in the inference engine. 

In this work, we focus on EP, a commonly used approximate inference
method. Following \cite{Heess2013}, we propose to learn a kernel-based
message operator for EP to capture the relationship between incoming
messages to a factor and outgoing messages. The operator bridges the
gap between the uninformed and informed cases by automatically deriving
the relevant computations for any custom factor that can be sampled.
This hybrid approach gives the modeler as much flexibility as in the
uninformed case while offering efficient message updates as in the
informed case. This approach supports fast inference as no expensive
KL divergence minimization needs to be done during inference as in
ordinary EP. In addition, a learned operator for a factor is reusable
in other models in which the factor appears. As will be seen, to send
an outgoing message with the kernel-based operator, it is sufficient
to generate a feature vector for incoming messages and multiply with
a pre-learned matrix. Unlike \cite{Heess2013} which considers a neural
network, the kernel-based message operator we propose can be easily
extended to allow online updates of the operator during inference. 

\begin{comment}
In the followings, we give a short introduction to EP in Sec. \ref{sec:Expectation-Propagation-(EP)},
and propose the kernel-based operator in Sec. \ref{sec:Learning-to-Pass}.
A proof-of-concept experiment on the operator is given in Sec. \ref{sec:Experiments}.
Finally, we conclude with remarks on future directions in Sec. \ref{sec:Conclusions-and-Future}.
\end{comment}

\begin{comment}
\begin{itemize}
\item the need for automated inference. =\textgreater{} learning a message
operator to send an outgoing message given incoming messages.
\item reusable, online update. Potential use in a probabilistic programming
language (PPL). 
\item Talk about a few PPLs. Talk about their approaches which are mostly
sampling. Drawback of sampling ?
\item Focus on EP here. A few comments on EP advantage ?
\end{itemize}
Split the conventional inference into 2 stages: training of a message
operator and actual inference.

Merits:

\textbf{Support fast inference. }No need to minimize multidimensional
KL divergence during actual inference. To send a message, compute
matrix $\times$ vector. \textbf{Automated inference}. No need to
derive complex message updates. \textbf{Applicable to any complex
factor. }Only requirement = ability to sample from the factor. No
need to be able to evaluate the likelihood value. Potential use in
a probabilistic programming language.\textbf{ Operator reusable }once
trained.
\end{comment}

\section{Expectation Propagation (EP)\label{sec:Expectation-Propagation-(EP)}}

Expectation propagation \citep{Minka2001,Bishop2006} (EP) is a commonly
used approximate inference method for inferring the posterior distribution
of latent variables given observations. In a typical directed graphical
model, the joint distribution of the data $X=\left\{ X_{1},\ldots,X_{n}\right\} $
and latent variables $\theta=\left\{ \theta_{1},\ldots,\theta_{t}\right\} $
takes the form of a product of factors, $p(X,\theta)=\prod_{i=1}^{m}f_{i}(X|\theta)$
where each factor $f_{i}$ may depend on only a subset of $X$ and
$\theta$. With $X$ observed, EP approximates the posterior with
$q(\theta)\propto\prod_{i=1}^{m}m_{f_{i}\rightarrow\theta}(\theta)$
where $m_{f_{i}\rightarrow\theta}$ is an approximate factor corresponding
to $f_{i}$ with the constraint that it has a chosen parametric form
(e.g., Gaussian) in the exponential family (ExpFam). EP takes into
account the fact that the final quantity of interest is the posterior
$q(\theta)$ which is given by the product of all approximate factors.
In finding the $i^{th}$ approximate factor $m_{f_{i}\rightarrow\theta}$,
EP uses other approximate factors $m_{\theta\rightarrow f_{i}}(\theta):=\prod_{j\neq i}m_{f_{j}\rightarrow\theta}(\theta)$
as a context to determine the plausible range of $\theta$. EP iteratively
refines $m_{f_{i}\rightarrow\theta}$ for each $i$ with $m_{f_{i}\rightarrow\theta}(\theta)=\frac{\text{proj}\left[\int dX\, f(X|\theta)m_{X\rightarrow f_{i}}(X)m_{\theta\rightarrow f_{i}}(\theta)\right]}{m_{\theta\rightarrow f_{i}}(\theta)}$
where $\text{proj}\left[r\right]=\arg\min_{q\in\text{ExpFam}}\text{KL}\left[r\,\|\, q\right]$
and $m_{X\rightarrow f_{i}}(X):=\delta(X-X_{0})$ if $X$ is observed
to be $X_{0}$. In the EP literature, $m_{\theta\rightarrow f_{i}}$
is known as a cavity distribution\textbf{. }

The projection can be carried out by the following moment matching
procedure. Assume an ExpFam distribution $q(\theta|\eta)=h(\theta)\exp\left(\eta^{\top}u(\theta)-A(\eta)\right)$
where $u(\theta)$ is the sufficient statistic\textbf{ }of $q$, $\eta$
is the natural parameter and $A(\eta)=\log\int d\theta\, h(\theta)\exp\left(\eta^{\top}u(\theta)\right)$
is the log-partition function. It can be shown that $q^{*}=\text{proj}\left[r\right]$
satisfies $\mathbb{E}_{q^{*}(\theta)}\left[u(\theta)\right]=\mathbb{E}_{r(\theta)}\left[u(\theta)\right].$
That is, the projection of $r$ onto ExpFam is given by $q^{*}\in\text{ExpFam}$
that has the same moment parameters as the moments under $r$. 

In general, under the approximation that each factor fully factorizes,
an EP message from a factor $f$ to a variable $V$ takes the form
\begin{equation}
m_{f\rightarrow V}(v)=\frac{\text{proj}\left[\int d\mathcal{V}\backslash\{v\}\, f(\mathcal{V})\prod_{V'\in\mathcal{V}}m_{V'\rightarrow f}(v')\right]}{m_{V\rightarrow f}(v)}:=\frac{\text{proj}\left[r_{f\rightarrow V}(v)\right]}{m_{V\rightarrow f}(v)}:=\frac{q_{f\rightarrow V}(v)}{m_{V\rightarrow f}(v)}\label{eq:general_ep_msg}
\end{equation}
where $\mathcal{V}=\mathcal{V}(f)$ is the set of variables connected
to $f$ in the factor graph. In the previous case of $m_{f_{i}\rightarrow\theta}$,
we have $\mathcal{V}(f)=\left\{ X,\theta\right\} $ and $V$ in Eq.
\ref{eq:general_ep_msg} corresponds to $\theta$. Typically, when
the factor $f$ is complicated, the integral defining $r_{f\rightarrow V}$
becomes intractable. Quadrature rules or other numerical integration
techniques are often applied to approximate the integral.

\section{Learning to Pass EP Messages\label{sec:Learning-to-Pass}}

Our goal is to learn a message operator $C_{f\rightarrow V'}$ with
signature $\left[m_{V\rightarrow f}\right]_{V\in\mathcal{V}(f)}\mapsto q_{f\rightarrow V'}$
which takes in all incoming messages $\left\{ m_{V\rightarrow f}\mid V\in\mathcal{V}(f)\right\} $
and outputs $q_{f\rightarrow V'}(v')$ i.e., the numerator of Eq.
\ref{eq:general_ep_msg}. For inference, we require one such operator
for each recipient variable $V'\in\mathcal{V}(f)$ i.e., in total
$|\mathcal{V}(f)|$ operators need to be learned for $f$. Operator
learning is cast as a distribution-to-distribution regression problem
where the training set $S_{V'}:=\{([m_{V\rightarrow f}^{n}]_{V\in\mathcal{V}(f)},q_{f\rightarrow V'}^{n})\}_{n=1}^{N}$containing
$N$ incoming-outgoing message pairs can be generated as in \cite{Heess2013}
by importance sampling to compute the mean parameters $\mathbb{E}_{r_{f\rightarrow V'}(v')}\left[u(v')\right]$
for moment matching. In principle, the importance sampling itself
can be used in EP for computing outgoing messages \citep{Barthelme2011}.
The scheme is, however, expensive as we need to draw a large number
of samples for each outgoing message to be sent. In our case, the
importance sampling is used for data set generation which is done
offline before the actual inference.

The assumptions needed for the generation of a training set are as
following. Firstly, we assume the factor $f$ takes the form of a
conditional distribution $f(v_{1}|v_{2},\ldots,v_{|\mathcal{V}(f)|})$.
Secondly, given $v_{2},\ldots,v_{|\mathcal{V}(f)|}$, $v_{1}$ can
be sampled from $f(\cdot|v_{2},\ldots,v_{|\mathcal{V}(f)|})$. The
ability to evaluate $f$ is not assumed. Finally we assume that a
distribution on the natural parameters of all incoming messages $\left\{ m_{V\rightarrow f}\right\} _{V\in\mathcal{V}(f)}$
is available. The distribution is used solely to give a rough idea
of incoming messages the learned operator will encounter during the
actual EP inference. In practice, we only need to ensure that the
distribution sufficiently covers the relevant region in the space
of incoming messages. 

In recent years, there have been a number of works on the regression
task with distributional inputs, including \cite{Poczos2013,Szabo2014}
which mainly focus on the non-parametric case and are operated under
the assumption that the samples from the distributions are observed
but not the distributions themselves. In our case, the distributions
(messages) are directly observed. Moreover, since the distributions
are in ExpFam, they can be characterized by a finite-dimensional natural
parameter vector or expected sufficient statistic. Hence, we can simplify
our task to distribution-to-vector regression where the output vector
contains a finite number of moments sufficient to characterize $q_{f\rightarrow V'}$.
As regression input distributions are in ExpFam, one can also treat
the task as vector-to-vector regression. However, seeing the inputs
as distributions allows one to use kernels on distributions which
are invariant to parametrization.

Once the training set $S_{V'}$ is obtained, any distribution-to-vector
regression function can be applied to learn a message operator $C_{f\rightarrow V'}$.
Given incoming messages, the operator outputs $q_{f\rightarrow V'}$
from which the outgoing EP message is given by $m_{f\rightarrow V'}=q_{f\rightarrow V'}/m_{V'\rightarrow f}$
which can be computed analytically. We opt for kernel ridge regression
\citep{Scholkopf2002} as our message operator for its simplicity,
its potential use in an online setting (i.e., incremental learning
during inference), and rich supporting theory.

\subsection{Kernel Ridge Regression}

We consider here the problem of regressing smoothly from distribution-valued
inputs to feature-valued outputs. We follow the regression framework
of \cite{Micchelli2005}, with convergence guarantees provided by
\cite{Caponnetto2007}. Under smoothness constraints, this regression
can be interpreted as computing the conditional expectation of the
output features given the inputs \citep{Grunewalder2012}.

Let $\mathsf{X}=\left(\mathsf{x}_{1}|\cdots|\mathsf{x}_{N}\right)$
be the training regression inputs and $\mathsf{Y}=\left(\mathbb{E}_{q_{f\rightarrow V'}^{1}}u(v')|\cdots|\mathbb{E}_{q_{f\rightarrow V'}^{N}}u(v')\right)\in\mathbb{R}^{D_{y}\times N}$
be the regression outputs. The ridge regression in the primal form
seeks $\mathsf{W}\in\mathbb{R}^{D_{y}\times D}$ for the regression
function $g(\mathsf{x})=\mathsf{W}\mathsf{x}$ which minimizes the
squared-loss function $J(\mathsf{W})=\sum_{i=1}^{N}\|\mathsf{y}_{i}-\mathsf{W}\mathsf{x}_{i}\|_{2}^{2}+\lambda\trace\left(\mathsf{W}\mathsf{W}^{\top}\right)$
where $\lambda$ is a regularization parameter and $\trace$ denotes
a matrix trace. It is well known that the solution is given by $\mathsf{W}=\mathsf{Y}\mathsf{X}^{\top}\left(\mathsf{X}\mathsf{X}^{\top}+\lambda I\right)^{-1}$
which has an equivalent dual solution $\mathsf{W}=\mathsf{Y}\left(K+\lambda I\right)^{-1}\mathsf{X}^{\top}=A\mathsf{X}^{\top}$
. The dual formulation allows one to regress from any type of input
objects if a kernel can be defined. All the inputs enter to the regression
function through the gram matrix $K\in\mathbb{R}^{N\times N}$ where
$\left(K\right)_{ij}=\kappa(\mathsf{x}_{i},\mathsf{x}_{j})$ yielding
the regression function of the form $g(\mathsf{x})=\sum_{i=1}^{N}a_{i}\kappa(\mathsf{x}_{i},\mathsf{x})$
where $A:=(a_{1}|\cdots|a_{N})$. The dual formulation therefore allows
one to straightforwardly regress from incoming messages to vectors
of mean parameters. Although this property is appealing, the training
size $N$ in our setting can be chosen to be arbitrarily large, making
computation of $g(\mathsf{x})$ expensive for a new unseen point $\mathsf{x}$.
To eliminate the dependency on $N$, we propose to apply random Fourier
features \citep{Rahimi2007} $\hat{\phi}(\mathsf{x})\in\mathbb{R}^{D}$
for $\mathsf{x}:=[m_{V\rightarrow f}]_{V\in\mathcal{V}(f)}$ such
that $\kappa(\mathsf{x},\mathsf{x}')\approx\hat{\phi}(\mathsf{x})^{\top}\hat{\phi}(\mathsf{x}')$
where $D$ is the number of random features. The use of the random
features allows us to go back to the primal form of which the regression
function $g(\hat{\phi}(\mathsf{x}))=\mathsf{W}\hat{\phi}(\mathsf{x})$
can be computed efficiently. In effect, computing an EP outgoing message
requires nothing more than a multiplication of a matrix $\mathsf{W}$
($D_{y}\times D$ ) with the $D$-dimensional feature vector generated
from the incoming messages.

\subsection{Kernels on Distributions}

A number of kernels on distributions have been studied in the literature
\citep{Jebara2003,Jebara2004}. Relevant to us are kernels whose random
features can be efficiently computed. Due to the space constraint,
we only give a few examples here.

\paragraph{Expected Product Kernel}

Let $\mu_{r^{(l)}}:=\mathbb{E}_{r^{(l)}(a)}k(\cdot,a)$ be the mean
embedding \citep{Smola2007} of the distribution $r^{(l)}$ into RKHS
$\mathcal{H}^{(l)}$ induced by the kernel $k$. Assume $k=k_{\text{gauss}}$
(Gaussian kernel) and assume there are $c$ incoming messages $\mathsf{x}:=(r^{(i)}(a^{(i)}))_{i=1}^{c}$
and $\mathsf{y}:=(s^{(i)}(b^{(i)}))_{i=1}^{c}$. The expected product
kernel $\kappa_{\text{pro}}$ is defined as 
\begin{align*}
\kappa_{\text{pro}}\left(\mathsf{x},\mathsf{y}\right) & :=\left\langle \bigotimes_{l=1}^{c}\mu_{r^{(l)}},\bigotimes_{l=1}^{c}\mu_{s^{(l)}}\right\rangle _{\otimes_{l}\mathcal{H}^{(l)}}=\prod_{l=1}^{c}\mathbb{E}_{r^{(l)}(a)}\mathbb{E}_{s^{(l)}(b)}k_{\text{gauss}}^{(l)}\left(a,b\right)\approx\hat{\phi}(\mathsf{x})^{\top}\hat{\phi}(\mathsf{y})
\end{align*}
where $\hat{\phi}(\mathsf{x})^{\top}\hat{\phi}(\mathsf{y})=\prod_{l=1}^{c}\hat{\phi}^{(l)}(r^{(l)})^{\top}\hat{\phi}^{(l)}(s^{(l)})$.
The feature map $\hat{\phi}^{(l)}(r^{(l)})$ can be estimated by applying
the random Fourier features to $k_{\text{gauss }}^{(l)}$and taking
the expectations $\mathbb{E}_{r^{(l)}(a)}\mathbb{E}_{s^{(l)}(b)}$.
The final feature map is $\hat{\phi}(\mathsf{x})=\hat{\phi}^{(1)}(r^{(1)})\circledast\hat{\phi}^{(2)}(r^{(2)})\circledast\cdots\circledast\hat{\phi}^{(c)}(r^{(c)})\in\mathbb{R}^{d^{c}}$where
$\circledast$ denotes a Kronecker product and we assume that $\hat{\phi}^{(l)}\in\mathbb{R}^{d}$
for $l\in\{1,\ldots,c\}$.

\paragraph{Kernel on Joint Embeddings }

Another way to define a kernel on $\mathsf{x},\mathsf{y}$ is to mean-embed
both joint distributions $\mathsf{r}=\prod_{i=1}^{c}r^{(i)}$ and
$\mathsf{s}=\prod_{i=1}^{c}s^{(i)}$ and define the kernel to be $\kappa_{\text{joint}}(\mathsf{x},\mathsf{y}):=\left\langle \mu_{\mathsf{r}},\mu_{\mathsf{s}}\right\rangle _{\mathcal{G}}$
where $\mathcal{G}$ is an RKHS consisting of functions $g:\mathcal{X}^{(1)}\times\cdots\times\mathcal{X}^{(c)}\rightarrow\mathbb{R}$
and $\mathcal{X}^{(l)}$ denotes the domain of $r^{(l)}$ and $s^{(l)}$.
This kernel is equivalent to the expected product kernel on the joint
distributions.\vspace{-3mm}

\section{Experiment\label{sec:Experiments}}

\begin{figure}
\begin{centering}
\includegraphics[width=6.2cm]{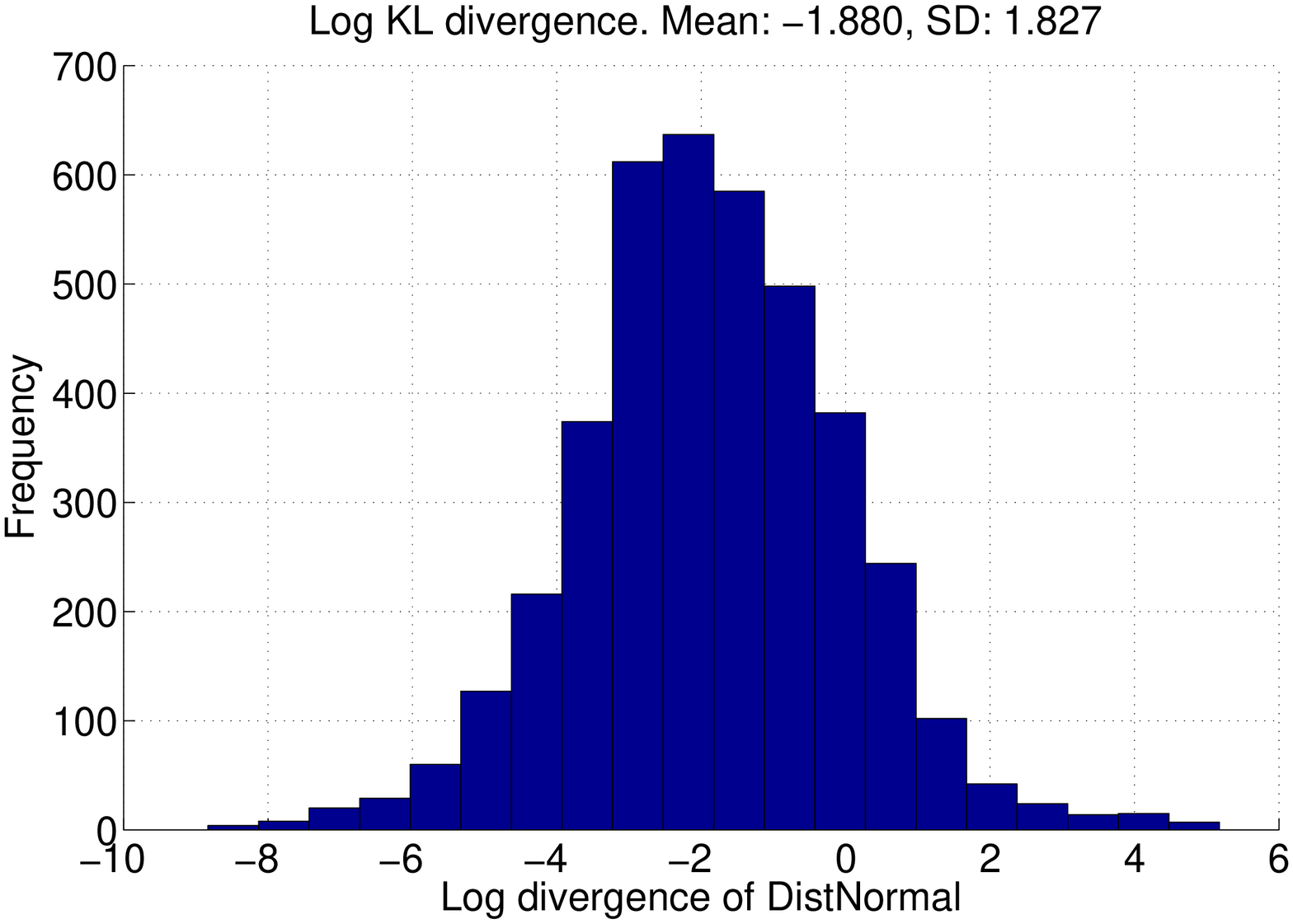}\includegraphics[width=6.5cm]{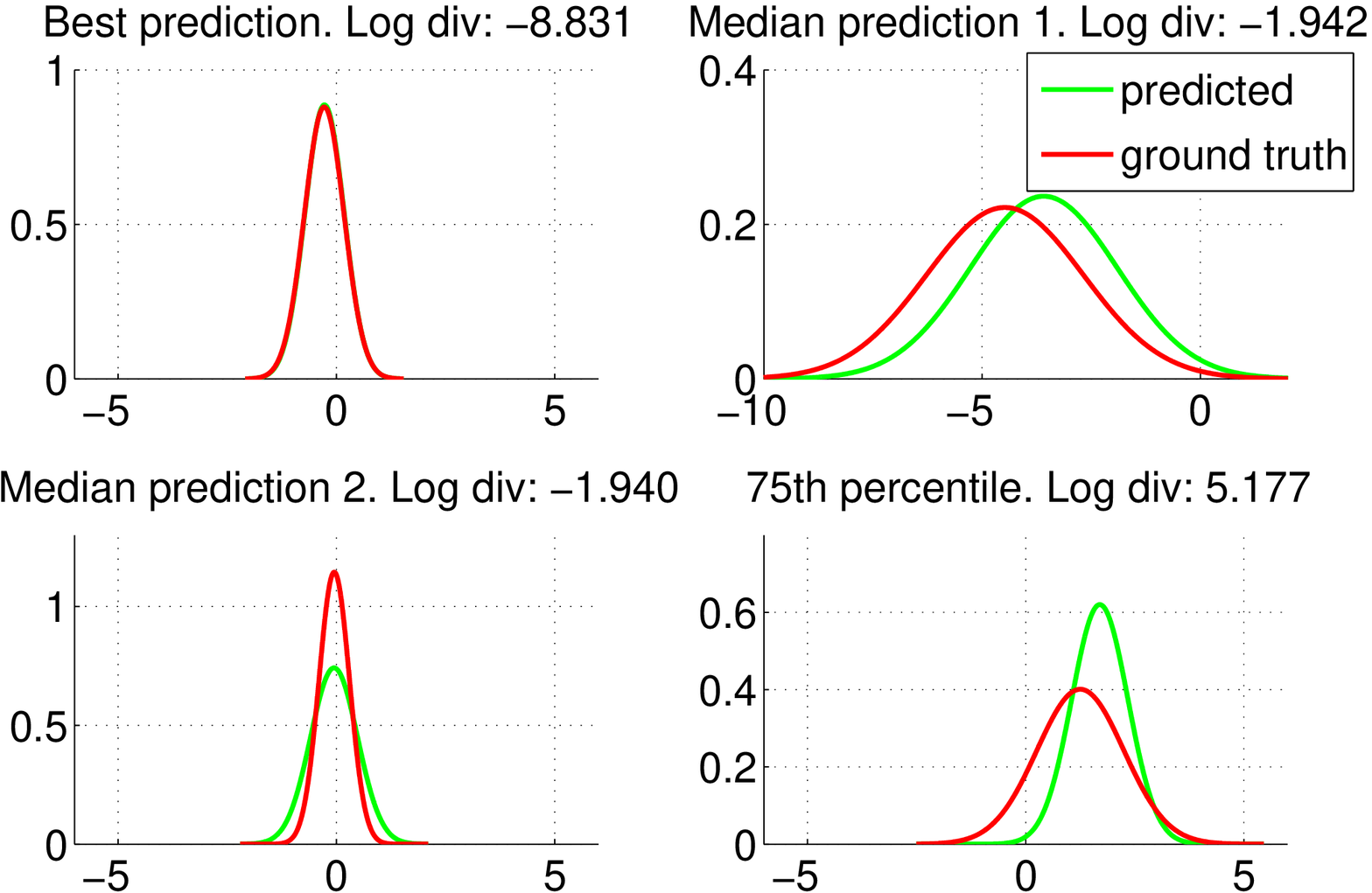}
\par\end{centering}

\begin{comment}
Joint embedding. exp1. RFGJointEProdLearner\_sigmoid\_bw\_proposal\_10000\_ntr2000.mat
\end{comment}
\vspace{-4mm}

\protect\caption{Log KL-divergence on a logistic factor test set using kernel on joint
embeddings.\label{fig:Log-KL-divergence-on}}
\end{figure}

\vspace{-3mm}As a preliminary experiment, we consider the logistic
factor $f(z|x)=\delta\left(z-\frac{1}{1+\exp(-x)}\right)$ which is
deterministic when conditioning on $x$. The factor is used in many
common models including binary logistic regression. The two incoming
messages are $m_{X\rightarrow f}(x)=\mathcal{N}(x;\mu,\sigma^{2})$
and $m_{Z\rightarrow f}(z)=\text{Beta}(z;\alpha,\beta)$. We randomly
generate 2000 training input messages and learn a message operator
using the kernel on joint embeddings. Kernel parameters are chosen
by cross validation and the number of random features $D$ is set
to 2000. We report $\log KL[q\|\hat{q}]$ where $q=q_{f\rightarrow X}$
is the ground truth output message obtained by importance sampling
and $\hat{q}$ is the message output from the operator. For better
numerical scaling, regression outputs are set to $(\mathbb{E}_{q}\left[x\right],\log\mathbb{V}_{q}\left[x\right])$
instead of the expectations of the first two moments. The histogram
of log KL errors is shown on the left of Fig. \ref{fig:Log-KL-divergence-on}.
The right figure shows sample output messages at different log KL
errors. It can be seen that the operator is able to capture the relationship
of incoming and outgoing messages. With higher training size, increased
number of random features and well chosen kernel parameters, we expect
to see a significant improvement in the operator's accuracy. \vspace{-3mm} 

\section{Conclusion and Future Work\label{sec:Conclusions-and-Future} }

\vspace{-2mm}We propose to learn to send EP messages with kernel
ridge regression by casting the KL minimization problem as a supervised
learning problem. With random features, incoming messages to a learned
operator are converted to a finite-dimensional vector. Computing an
outgoing message amounts to computing the moment parameters by multiplying
the vector with a matrix given by the solution of the primal ridge
regression.

By virtue of the primal form, it is straightforward to derive an update
equation for an online-active learning during EP inference: if the
predictive variance (similar to a Gaussian process) on the current
incoming messages is high, then we query the outgoing message from
the importance sampler (oracle) and update the operator. Otherwise,
the outgoing message is efficiently computed by the operator. Online
learning of the operator lessens the need of the distribution on natural
parameters of incoming messages used in training set generation. Determining
an appropriate distribution was one of the unsolved problems in \cite{Heess2013}.

\subsubsection*{Acknowledgments }

We gratefully acknowledge the support of the Gatsby Charitable Foundation. 

\newpage\bibliographystyle{abbrvnat}
\bibliography{ref}

\end{document}